\documentclass[preprint,12pt,3p]{elsarticle}

\usepackage{lineno,hyperref}
\usepackage{color}
\usepackage{multirow}
\usepackage{amsmath,amssymb,amsfonts}
\usepackage[ruled]{algorithm2e}
\usepackage{graphicx}
\usepackage{textcomp}
\usepackage{subfigure}
\usepackage[misc]{ifsym}
\usepackage{booktabs}
\usepackage{times}
\usepackage{footnote}

\journal{Neural Networks}

\begin{document}
	
	\begin{frontmatter}
		
		\title{Multi-Granularity Graph Pooling for Video-based Person Re-Identification}
		
		\author[mymainaddress]{Honghu Pan}
		\author[mymainaddress]{Yongyong Chen}
		\author[mymainaddress]{Zhenyu He\corref{mycorrespondingauthor}}
		\ead{zhenyuhe@hit.edu.cn}
		
		\cortext[mycorrespondingauthor]{Corresponding author}
		\address[mymainaddress]{School of Computer Science and Technology, Harbin Institute of Technology, Shenzhen 518055, China}

		\begin{abstract}
		The video-based person re-identification (ReID) aims to identify the given pedestrian video sequence across multiple non-overlapping cameras.
		To aggregate the temporal and spatial features of the video samples, the graph neural networks (GNNs) are introduced.
		However, existing graph-based models, like STGCN, perform the \textit{mean}/\textit{max pooling} on node features to obtain the graph representation, which neglect the graph topology and node importance.
		In this paper, we propose the graph pooling network (GPNet) to learn the multi-granularity graph representation for the video retrieval, where the \textit{graph pooling layer} is implemented to downsample the graph.
		We first construct a multi-granular graph, whose node features denote image embedding learned by backbone, and edges are established between the temporal and Euclidean neighborhood nodes.
		We then implement multiple graph convolutional layers to perform the neighborhood aggregation on the graphs.
		To downsample the graph, we propose a multi-head full attention graph pooling (MHFAPool) layer, which integrates the advantages of existing node clustering and node selection pooling methods.
		Specifically, MHFAPool takes the main eigenvector of full attention matrix as the aggregation coefficients to involve the global graph information in each pooled nodes.
		Extensive experiments demonstrate that our GPNet achieves the competitive results on four widely-used datasets, i.e., MARS, DukeMTMC-VideoReID, iLIDS-VID and PRID-2011.
		\end{abstract}
		
		
	\end{frontmatter}

\section{Introduction}
In the past several years, the image-based person re-identification (ReID)~\cite{FPO,EEA,PIGA,TriNet} has achieved great improvements thanks to the deep convolutional neural networks (CNNs).
However, the image-based person ReID would become very challenging when the occlusion or pose variation occurs.
As an important surveillance data, video contains the rich temporal information, therefore, a great number of studies~\cite{Mars,duke1,STGCN} have attempted the video sequences for pedestrian retrieval, which aims to mine the temporal correlations among multiple frames and the spatial relations of the pedestrian body parts.
Given a query sample with a specific identity, the video-based person ReID aims to find the video sequences of the same identity in the gallery video set.

The current video-based ReID methods mainly learn a neural network that converts the input video sequence into a representation vector.
Generally, the learning procedure can be summarized as two steps: feature extraction by a backbone and feature aggregation by an aggregator.
As the image-based ReID, the CNNs are served as the backbone to extract the feature maps from video frames.
The crux of the video-based ReID is to design an aggregator for the feature aggregation with fully considering the temporal and spatial clues contained in the video sequences: the temporal clues primarily indicate the continuous motion and posture information of the pedestrian, while the spatial clues denote the correlations among the part-level features.

The graph neural networks (GNNs)~\cite{GraphSAGE,MPNN,GCN}, which are developed for the feature learning of the non-Euclidean data, match the demands of aggregator well.
Specifically, the GNNs perform the neighborhood aggregation on graph data with the graph convolution operation, thereby it could be utilized to the feature aggregation in video-based ReID.
However, the GNNs propagate the graph information within the fixed graph structure, in other words, the number of nodes and edges is constant after the graph convolutional layers.
This brings the obstacles to obtaining the feature representation of the input graph.
Existing graph-based works, like STGCN~\cite{STGCN}, perform the \textit{mean}/\textit{max pooling} on node features, which neglect the graph topology and the node importance.
More specifically, the \textit{mean}/\textit{max pooling} ignores the adjacent relations between nodes, and treats the important nodes (i.e., pedestrian body features) and the unimportant nodes (e.g., occlusion or background) equally.

To resolve this issue, we propose the graph pooling network, a multi-branch architecture to learn the multi-granularity graph representations, where the \textit{graph pooling layer} is implemented to downsample the graph.
To fully exploit the temporal and spatial clues, the multi-granularity features are employed as the node features to construct several graphs, where each graph corresponds to a specific granularity. The connections or edges are established between the temporal and Euclidean neighborhood nodes. With the graph convolution layers of GNNs, each node aggregates information from its adjacent nodes, which enables the temporal and spatial aggregation of the video-based ReID.

\begin{figure}[t]
	\centering
	\subfigure[Node clustering]{\includegraphics[width=0.31\textwidth]{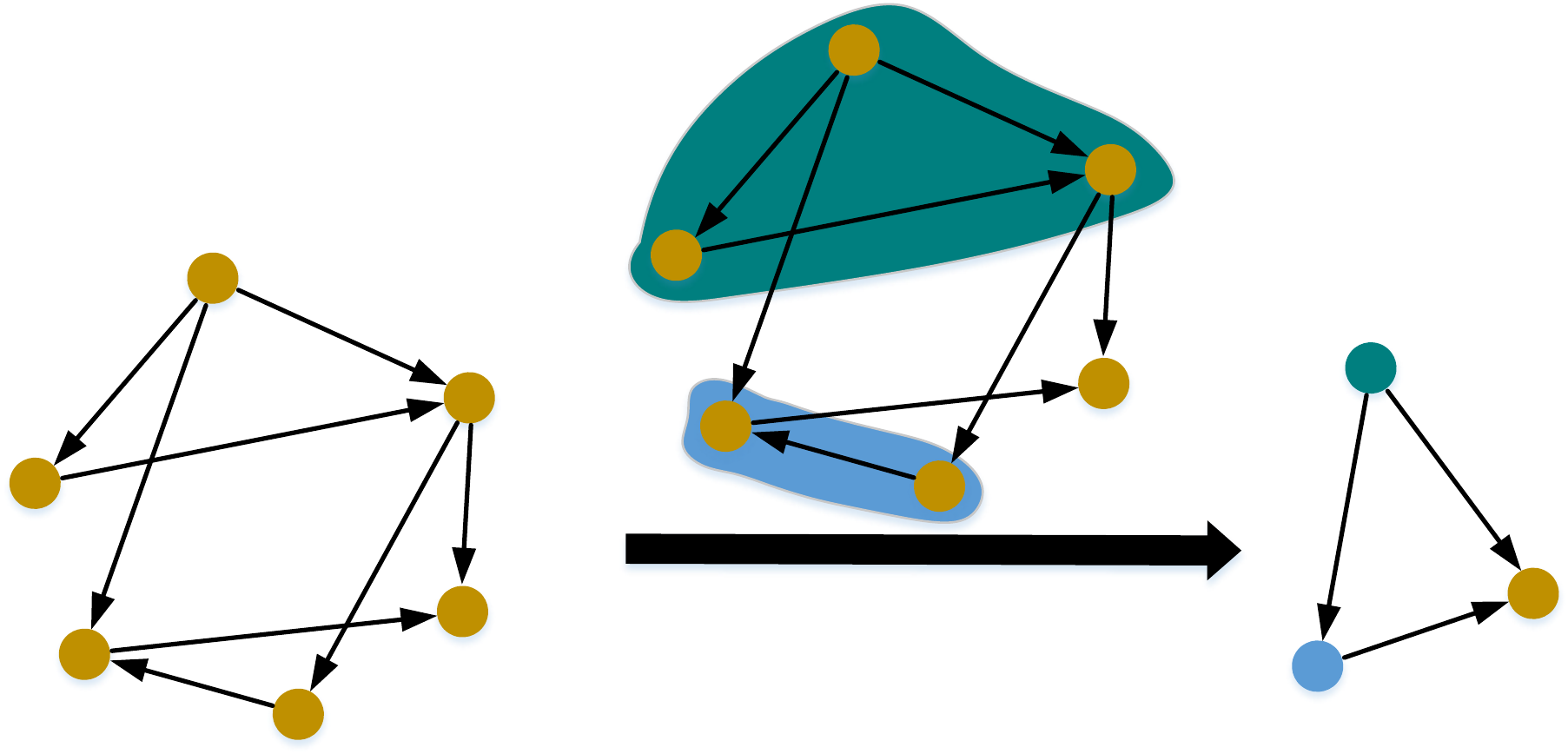}}
	\subfigure[Node selection]{\includegraphics[width=0.35\textwidth]{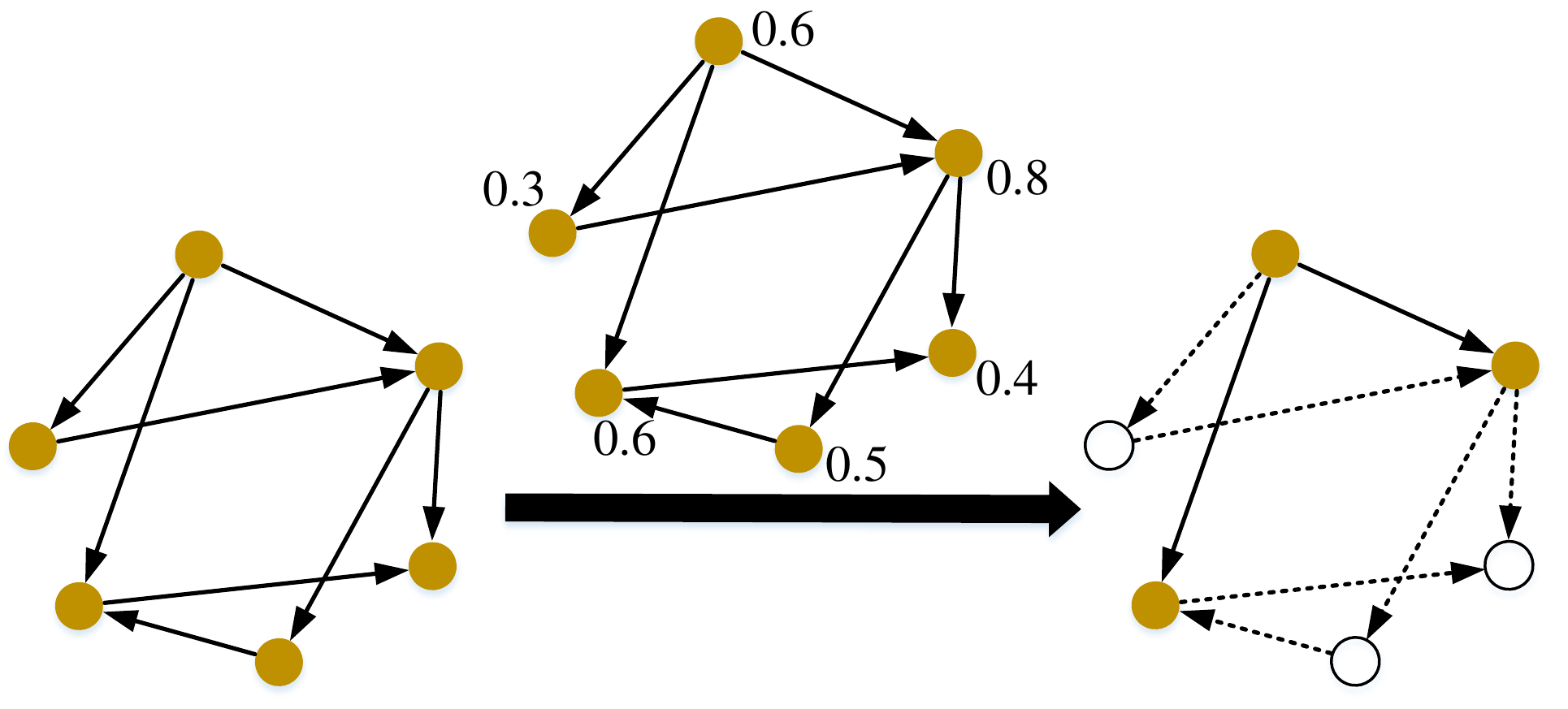}}
	\caption{The existing graph pooling methods. (a) The node clustering-based method performs the graph coarsening by clustering the adjacent nodes. (b) The node selection-based method retains the nodes with top-$k$ scores and discards the remaining nodes.}
	\label{fig_pooling}
\end{figure}

To learn the graph representation, we introduce the \textit{graph pooling}~\cite{DiffPool,SAGPool,eigenpooling}, an important theory in machine learning, to scale down the input graph with decreasing the number of nodes.
As shown in Fig.~\ref{fig_pooling}, existing graph pooling methods can be categorized as the \textit{node clustering-based} and \textit{node selection-based}, and their representative algorithms are DiffPool~\cite{DiffPool} and SAGPool~\cite{SAGPool}, respectively.
Between them, DiffPool performs the soft assignment clustering on the adjacent nodes, while SAGPool learns the node scores with the self-attention mechanism, and then retains the nodes with top-$k$ scores.
However, both DiffPool and SAGPool have its own disadvantage:
DiffPool only performs the local node clustering and neglects node importance, while SAGPool discards part of the graph information.
Thereby, this paper proposes a multi-head full attention pooling (MHFAPool) method, by which every pooled node contains the global graph information.
Specifically, for each pooled node, we first compute a full attention square matrix, then employ the \textbf{power iteration algorithm} to quickly calculate the main eigenvector of the square matrix, and last derive the aggregation coefficients via Softmax operation.
The naive motivation of MHFAPool is that the main eigenvector contains the richest information of a matrix.
And the multi-head structure allows us to retain multiple nodes after graph pooling.
Besides the pooling layer, we set a readout layer as the last layer to obtain the fixed length representation of the pooled graph.

We dub the proposed graph pooling network as the GPNet.
In our GPNet, the feature aggregation is divided into two main steps: 1) feature neighborhood aggregation by the graph convolutional layers; 2) graph representation learning by the graph pooling and readout layer.
With the multi-granularity graph representations, we concatenate them together to form the final representation vector of the input video sequence.
We test our GPNet on four widely-used datasets, i.e., MARS~\cite{Mars}, DukeMTMC-VideoReID~\cite{duke1,duke2}, iLIDS-VID~\cite{ilids} and PRID-2011~\cite{prid}, and the experimental results demonstrate that the graph pooling methods show a significant improvement over the mean/max pooling.
The main contributions of this paper are three-fold:
\begin{itemize}
	\item We propose the multi-granularity graph aggregation by GNNs to capture the temporal and spatial clues, where the graphs are constructed by the multi-granularity features of the video sequence;
	\item We propose MHFAPool to downsample the graph via the attention mechanism and power iteration algorithm, which cold preserve the graph information to the greatest extent and meanwhile consider the node importance;
	\item Our model achieves the competitive results on four video-based ReID datasets, which validates the effectiveness of the proposed GPNet.
\end{itemize}

The remainder of this paper is organized as follows:
Section~\ref{Related_Works} introduces the related works, including person re-identification, graph neural networks and graph pooling methods;
Section~\ref{method} illustrates our GPNet in detail;
Section~\ref{experiments} reports the experimental results in multiple video-based person re-identification datasets;
Section~\ref{conclusion} presents the conclusions of this paper.

\section{Related Works}
\label{Related_Works}
\subsection{Image-based Person Re-Identification}
The image-based person ReID has been studied for several decades. 
In recent years, the deep features play a dominated role compared with the handcrafted descriptors.
Generally, a standard pipeline for the image-based person ReID is composed of two modules~\cite{reid_survey}: feature representation learning and deep metric learning.
The feature representation learning focuses on constructing feature vectors from the pedestrian images with the deep CNNs. 
The widely used feature representations include the global features~\cite{reranking,TriNet,BoT}, local features~\cite{MGN,PCB,suh2018part} and attribute features~\cite{attribute1,ADFD}.
Meanwhile, the multi-granularity features (global features + local features) can effectively improve the ReID accuracy, which are also adopted in this paper.
The deep metric learning defines the training objectives with the loss functions and sampling strategies.
With respect to the loss functions, the identity loss, triplet loss~\cite{facenet} and center loss~\cite{centerloss} have been successfully applied in the image-based ReID~\cite{BoT,centerlearning}.
For the sampling strategies, the hard triplet mining~\cite{TriNet} and adaptive sampling~\cite{curriculumsampling} are popular approaches.

\subsection{Video-based Person Re-Identification}
Beyond the spatial correlation in a single image, video, which contains more temporal information, has been extended to the person ReID.
A large number of methods have been proposed to handle the video-based person ReID task.
For instance, the optical flow-based models~\cite{optical1,optical2} are employed to learn the short-term temporal dependency, while the 3D convolution~\cite{3DCNN1,3DCNN2} and the RNN-based models~\cite{RNN1,RNN2} can capture the long-term temporal clues:
AP3D~\cite{3DCNN1} learned the pedestrian appearance information with an appearance-preserving module in the 3D convolution-based framework;
RCNet~\cite{RNN1} combined the recurrent layer and the temporal pooling layer to extract the feature of input videos.
In recent years, several studies took advantage of the part-level features to model both temporal and spatial clues with the attentive mechanism~\cite{COSAM,RAFA,NVAN} or graph model~\cite{MGH,STGCN}:
COSAM~\cite{COSAM} proposed a co-segmentation based attention module to activate the salient features;
MG-RAFA~\cite{RAFA} implemented the multi-granularity attentive feature aggregation for the video representation learning.
MGH~\cite{MGH} proposed the hyper-graph aggregation for the multi-granularity features, in which a hyper-edge connects multiple nodes;
STGCN~\cite{STGCN} applied the graph convolutional network (GCN) to model the spatial and temporal relations of the part-level features.
There exist two main differences between our GPNet and the STGCN:
STGCN constructed a temporal graph and a spatial graph to capture the temporal and spatial clues, while our GPNet deploys the multi-granularity features and constructs a graph for each granularity;
otherwise, STGCN adopted the max pooling to obtain the feature representation of a graph, while our GPNet introduces the graph pooling to decrease the number of nodes in graph.

\subsection{Graph Neural Networks and Graph Pooling}
As CNNs perform the local filtering on images, the GNNs define the localized graph convolution on the graph data to achieve the neighborhood aggregation mechanism.
The graph convolution can be divided into the spatial-based convolution and the spectral-based convolution.
The spatial-based graph convolution achieves the neighborhood aggregation with the spatial relations of nodes:
MPNN~\cite{MPNN} summarized the graph convolution as the message passing mechanism; GraphSAGE~\cite{GraphSAGE} enabled the inductive learning by the neighbor sampling.
The spectral-based graph convolution focuses on the global structure of input graph, and defines the graph convolution with the graph Laplacian matrix: 
the spectral network~\cite{GCN1} directly learned a graph filter in the graph Fourier domain;
ChebNet~\cite{GCN2} approximated the graph filter with the Chebyshev polynomials to achieve the localized convolution;
GCN~\cite{GCN} implemented the layer-wise architecture with the 1-st order Chebyshev polynomials to alleviate the overfitting.	

Graph pooling is a crucial ingredient in the graph-level tasks (e.g., graph classification), which aims to downsample the graph or decrease the number of nodes in graph.
As shown in Fig.~\ref{fig_pooling}, the current graph pooling methods can be divided into two categories: node clustering-based and node selection-based.
The node clustering-based method performs the graph coarsening with a learning-based assignment matrix:	
DiffPool~\cite{DiffPool} proposed to learn the assignment matrix with a separate GNN branch to enable the differentiable soft clustering;
EigenPooling~\cite{eigenpooling} utilized the eigenvector of the graph Laplacian matrix to define the assignment matrix;
MinCutPool~\cite{MinCutPool} clustered and aggregated nodes with the spectral clustering technique.
The node selection-based method retains the nodes with top-$k$ scores and discards the remaining nodes:
Graph U-Net~\cite{graphU} learned the scores from the node feature vectors;
SAGPool~\cite{SAGPool} evaluated the node scores by the self-attention mechanism.

\section{Methodology}
\label{method}
In this section, we first present the overview of our model in Section~\ref{Overview}, then illustrate the GPNet in Section~\ref{GPNet}, last give the loss functions for model training and inference pattern in Section~\ref{loss}.

\subsection{Problem Definition and Overview}
\label{Overview}
The representation learning is widely adopted in video-based person ReID.
Given a video sequence $V_I=\{I_1,I_2,\cdots,I_T\}$ with $T$ frames, which corresponds to a specific pedestrian identity $y_I$, we aim to learn a feature extractor to transform the pedestrian video $V_I$ into a representation vector.
The objective is to enlarge the the similarity of feature vectors corresponding to the same identity.

Normally, a CNN backbone is employed to extract the feature of each frame.
Thus, we can learn the feature map sequence $\mathbb{F}=\{\mathbb{F}_1,\mathbb{F}_2,\cdots,\mathbb{F}_T\}$ from the video sequence, where $\mathbb{F}_i \in R^{w \times h \times c}$ indicates the feature map of image $I_i$; $w$, $h$, $c$ denote the width, height and channel size, respectively.
With the feature map sequence $\mathbb{F}$, we then obtain the multi-granularity features (i.e. global features and part-level features): 
we perform the average pooling on $\mathbb{F}_i$ to construct the 1-st order granularity features (global features) $F^{global}=F_1=\{f_1^{(1)},f_2^{(1)},\cdots,f_T^{(1)}\}$, where $f_i^{(1)} \in R^{c \times 1}$;
we horizontally partition $\mathbb{F}_i$ into $p$ parts and then perform the average pooling on the partitioned features to construct the $p$-th order granularity features (part-level features) $F_p=\{f_1^{(1)},f_2^{(2)},\cdots,f_i^{(p-1)},\cdots,f_{T\times p}^{(p)}\}$, where the superscript indicates the part of pedestrian body as in Fig.~\ref{fig_adjacency}.
In this paper, we set $p$ as $\{1,2,4,8\}$.

\begin{figure*}[t]
	\centering
	\includegraphics[width=0.98\textwidth]{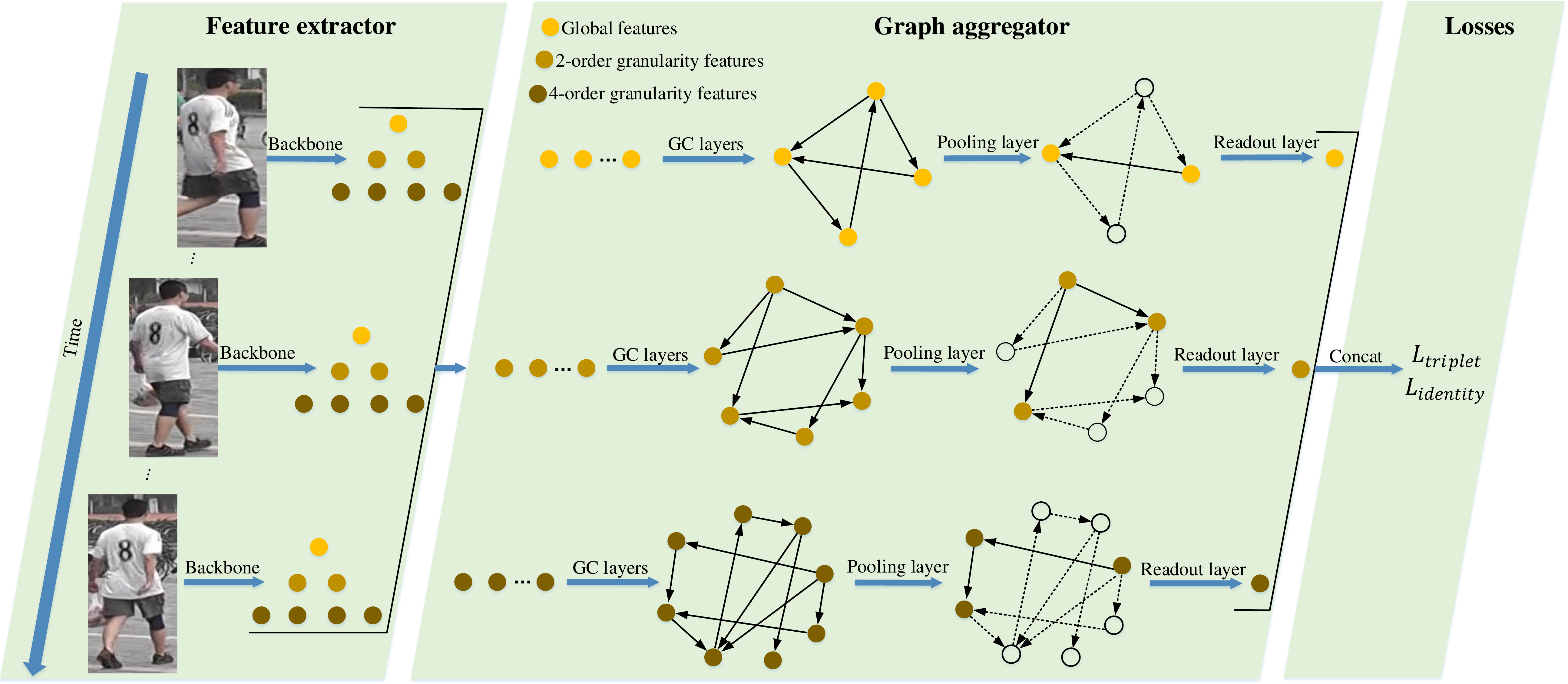}
	\caption{The overall architecture of the proposed GPNet.
		We first learn the multi-granularity features with the backbone, then construct several graphs with them, where each graph corresponds to a specific granularity.
		The GC layers perform the neighborhood aggregations on the graph to capture the temporal and spatial clues; the pooling layer downsamples the graph with the graph pooling operation; and the readout layer learns the final representation of graph.
		We concatenate the multi-granularity graph representations as the representation vector of the input video sequence. 
		We train our model with the triplet loss and identity loss.
	}
	\label{fig_model}
\end{figure*}

With the multi-granularity features, the video-based ReID aims to design an aggregator to perform the temporal and spatial aggregation from $F_p$ to the video representation.
In this paper, we propose the graph pooling network (GPNet) to aggregate the multi-granularity features.
As shown in Fig.~\ref{fig_model}, our GPNet contains multiple branches, and each of them corresponds to a specific granularity.
For each branch, we construct a graph with the feature of the corresponding granularity for the subsequent learning processes.
The GPNet divides the feature aggregation into two steps: the neighborhood aggregation by the graph convolutional (GC) layers and the graph representation learning by the pooling and readout layer.
In the next section we will elaborate our GPNet in detail.

\subsection{Graph Pooling Network (GPNet)}
\label{GPNet}
In this section, we first introduce how to construct the multi-granularity graphs, then elaborate the GC layer, pooling layer and readout layer in our GPNet, respectively.
For the pooling layer, we present three graph pooling methods, including DiffPool~\cite{DiffPool}, SAGPool~\cite{SAGPool}, and the MHFAPool proposed by this paper.

\subsubsection{Graph Construction}
A graph, which is composed of nodes and edges, can be used as the input of GNNs for the feature aggregation.	
To construct the input graphs of our model, the features learned by the backbone are taken as the node features.
Specifically, we construct a separate graph with the features of each granularity, so that we can formulate the multi-branch architecture with the constructed multiple graphs.
Therefore, the input graph of each branch contains a specific number of nodes:
for the global branch, the input graph contains $T$ nodes $v_1,v_2,\cdots,v_T$, whose features are $f_1^{(1)},f_2^{(1)},\cdots,f_T^{(1)}$;
for the part-level features, the graph contains $T \times p$ nodes, where $p$ denotes the granularity order.

\begin{figure}[t]
	\centering
	\includegraphics[width=0.48\textwidth]{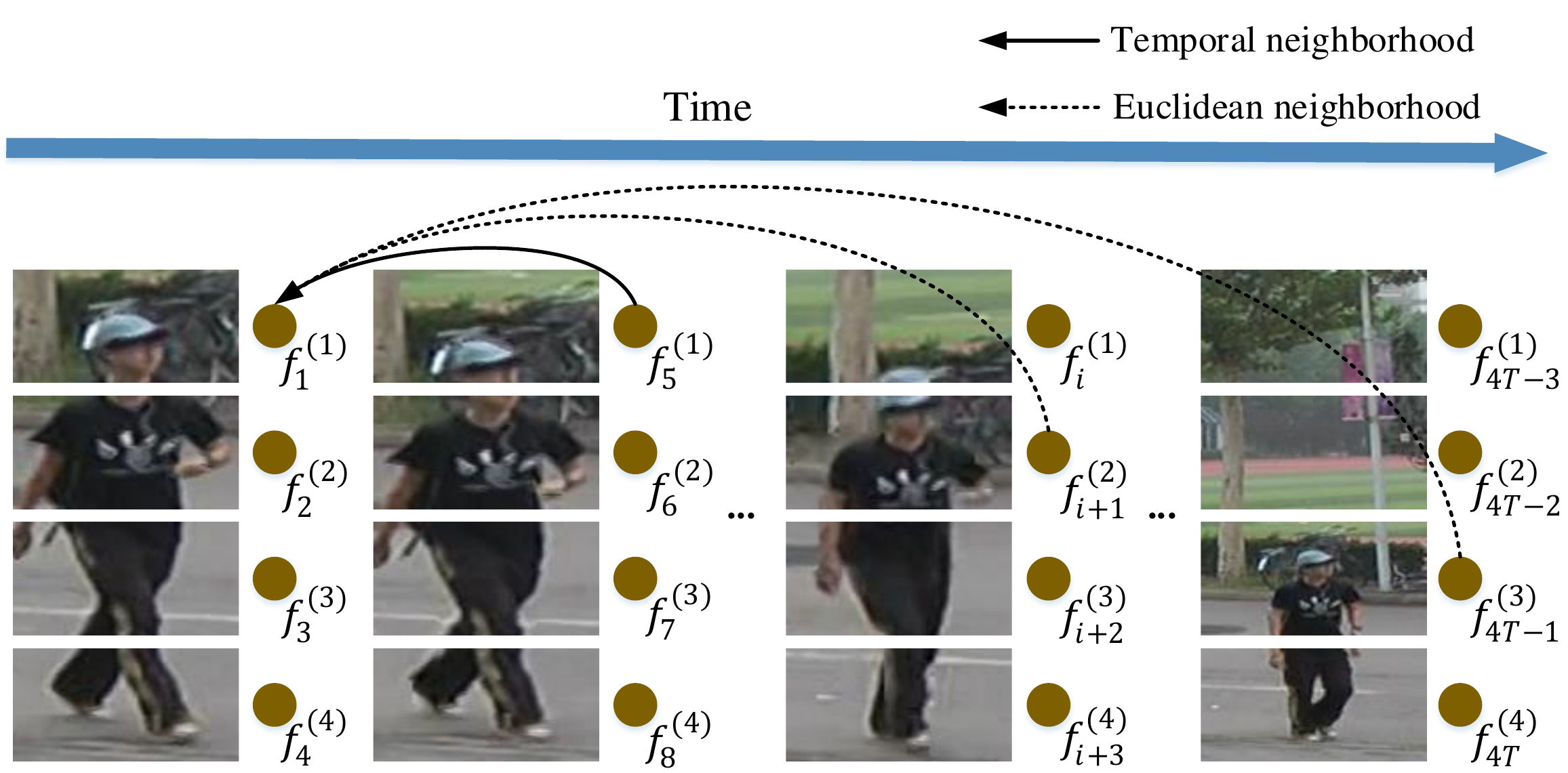}
	\caption{Temporal neighborhood aggregation and Euclidean neighborhood aggregation of feature $f_1^{(1)}$, where $f_5^{(1)}$ denotes the temporal neighborhood feature of $f_1^{(1)}$; $f_{i+1}^{(2)}$ and $f_{4T-1}^{(3)}$ denote the Euclidean neighborhood features of $f_1^{(1)}$.}
	\label{fig_adjacency}
\end{figure}

With the nodes in graph, we then define the edges to connect the neighborhood nodes.
To fully capture the temporal and spatial clues, we should aggregate both the temporal neighbors and spatial (or Euclidean) neighbors for each node.
To this end, we propose the dual-neighborhood aggregation (DNA) mechanism for the graph message passing:
we define $f_j^{(p_j)}$ is the temporal neighborhood feature of $f_i^{(p_i)}$ if 1) they correspond to the same part of the pedestrian images ($p_j=p_i$) and 2) their frames are temporal neighbors ($|i-j|\leq \delta_T$);
we define $f_j^{(p_j)}$ is the Euclidean neighborhood feature of $f_i^{(p_i)}$ if $f_j^{(p_j)}$ is in the $k$-nearest neighbor ($k$NN) set of $f_i^{(p_i)}$ under the Euclidean distance metric.
Mathematically, the established adjacent relations or edges in graph can be denoted as:
\begin{equation}
\begin{split}
e_{\vec{<ji>}}=\{ <v_j \rightarrow v_i>, \ & {\rm if} \ \{ p_j=p_i, \ |i-j|\leq \delta_T \} \\
&{\rm or} \ \{ f_j^{p_j} \in kNN (f_i^{p_i}) \}   \},
\label{eq_edge}
\end{split}
\end{equation}
where $f_i^{(p_i)}$ and $f_j^{(p_j)}$ denote the features of nodes $v_i$ and $v_j$ ,respectively, and $\delta_T$ is a hyper-parameter.
As shown in Fig.~\ref{fig_adjacency}, the temporal neighborhood aggregation aims to capture the temporal dependency of the features, while the Euclidean neighborhood aggregation can solve the misalignment of pedestrian parts due to the long-term displacement (e.g. $f_1^{(1)}$ and $f_{4T-1}^{(3)}$).

\subsubsection{GC Layers}
With the input graph, we then implement the GC layers of GNNs to perform the neighborhood aggregation for each node.
Generally, the graph convolution can be categorized as the spatial-based graph convolution and the spectral-based graph convolution: the former defines the neighborhood aggregation with the spatial relations of nodes, while the latter is developed from the graph spectral theory.
In this paper, we attempt both the spatial-based and the spectral-based graph convolution for the GC layers, and their respective performances are explored in the ablation studies of Section~\ref{Ablation_Studies}.

\textbf{Spatial-based Graph Convolution:}
We follow the convolution operation proposed in GraphSAGE~\cite{GraphSAGE}, by which, the spatial-based graph convolution summarizes the neighborhood aggregation as two steps, i.e. aggregating information from the adjacent nodes and updating node information:
\begin{equation}
h^{(l)}_{\mathcal{N}(v_i)} = \frac{1}{c_i} \sum_{j=1}^{c_i} h^{(l-1)}_{j}, \ \forall v_j \in \mathcal{N}(v_i),
\end{equation}
\begin{equation}
h^{(l)}_{i} = \sigma \{ w^{(l)}_1 h^{(l-1)}_{i} || w^{(l)}_2 h^{(l)}_{\mathcal{N}(v_i)} \},
\end{equation}
where $h^{(l)}_{i}$ denotes the feature of node $v_i$ in $l$-th layer; $h^{(0)}_{i}$ indicates the initial node features, i.e. feature learned by the backbone $f_i^{(p_i)}$; $v_j$ is the neighborhood node of $v_i$; $c_i$ denotes the number of neighborhood nodes of $v_i$; $ w^{(l)}_1$ and $w^{(l)}_2$ are the learnable parameters; $\sigma (\cdot)$ is the non-linear activation function, and $||$ represents the \textit{concat} operation.

\textbf{Spectral-based Graph Convolution:}
The spectral-based graph convolution~\cite{GCN} focuses on the structure of the whole graph, and defines the graph convolution with the graph Laplacian matrix, whose propagation rule can be expressed as follows:
\begin{equation}
H^{(l)} = \sigma (\tilde{D}^{-1/2} \tilde{A} \tilde{D}^{-1/2} H^{(l-1)} W^{(l)}) ,
\label{eq_GCN}
\end{equation}
where
\begin{equation}
\tilde{A}=A+I_n ,\\
\label{eq_adjacency1}
\end{equation}
and
\begin{equation}
A_{ij}=\left\{
\begin{aligned}
&1, \ \ {\rm if} \ \ e_{\vec{<j i>}}\\
&0, \ \ {\rm otherwise}.\\
\end{aligned}
\label{eq_adjacency2}
\right.
\end{equation}
In Eq.~(\ref{eq_adjacency1}), $I_n$ represents the identity matrix and $n$ denotes the number of nodes in the graph.
In Eq.~(\ref{eq_GCN}), $\tilde{D}_{ii} = \sum_{j} \tilde{A}_{ij}$ denotes the degree matrix; $H^{(l)}=\{h^{(l)}_{i}, i=1,2,\cdots,n \}$ denotes the node features in $l$-th layer; and $W^{(l)}$ is the learnable parameter.

With the graph convolution operation, we can implement the GC layers for the temporal-spatial aggregation of the ReID features.
After $l$ GC layers, each node contains the information of its $l$-hop neighborhood nodes.
By this non-linear aggregation, we can capture the long-term temporal dependency and mine the multi-granularity spatial clues.

\subsubsection{Graph Pooling Layer}
\label{pooling}
As pooling in CNNs can decrease the size of feature maps, the graph pooling reduces the number of nodes in graph with considering the graph topology.
In this paper, we first attempt the node clustering-based method and the node selection-based method for the pooling layer, whose representative algorithms are DiffPool~\cite{DiffPool} and SAGPool~\cite{SAGPool}, respectively.
And then we elaborate the novel multi-head full attention pooling (MHFAPool).
The performances of DiffPool~\cite{DiffPool}, SAGPool~\cite{SAGPool} and MHFAPool are explored in the ablation studies of Section~\ref{Ablation_Studies}.

\textbf{DiffPool-based Pooling.}
DiffPool first learns a soft assignment matrix $S$ with a separate GNN branch:
\begin{equation}
S = {\rm softmax} \{GNN_{pool} (A, H^{(GC)}) \},
\label{eq_S}
\end{equation}
and then performs the graph coarsening by clustering the original nodes with $S$:
\begin{equation}
H^{(P)} = S^T H^{(GC)}.
\label{eq_pooling1}
\end{equation}
In Eq.~(\ref{eq_S}) and Eq.~(\ref{eq_pooling1}), $H^{(GC)}$ denotes the node features learned by the GC layers: for the spatial-based convolution, $H^{(GC)}$ indicates the features of the last layer $H^{(l)}$; for the spectral-based convolution, $H^{(GC)}$ indicates the shortcut connection of node features in each layer, i.e. $H^{(GC)}=FC(H^{(1)} || H^{(2)} ||\cdots|| H^{(l)})$.
The size of the soft assignment matrix $S$ is equal to $n \times m$, therefore, the number of nodes would decrease from $n$ to $m$.
Compared to the mean/max pooling, DiffPool requires two extra steps: computation of assignment matrix $S$ via Eq.~(\ref{eq_S}) and node clustering via Eq.~(\ref{eq_pooling1}). This brings extra computation for matrix multiplication.

\begin{figure}[t]
	\centering
	\includegraphics[width=0.70\textwidth]{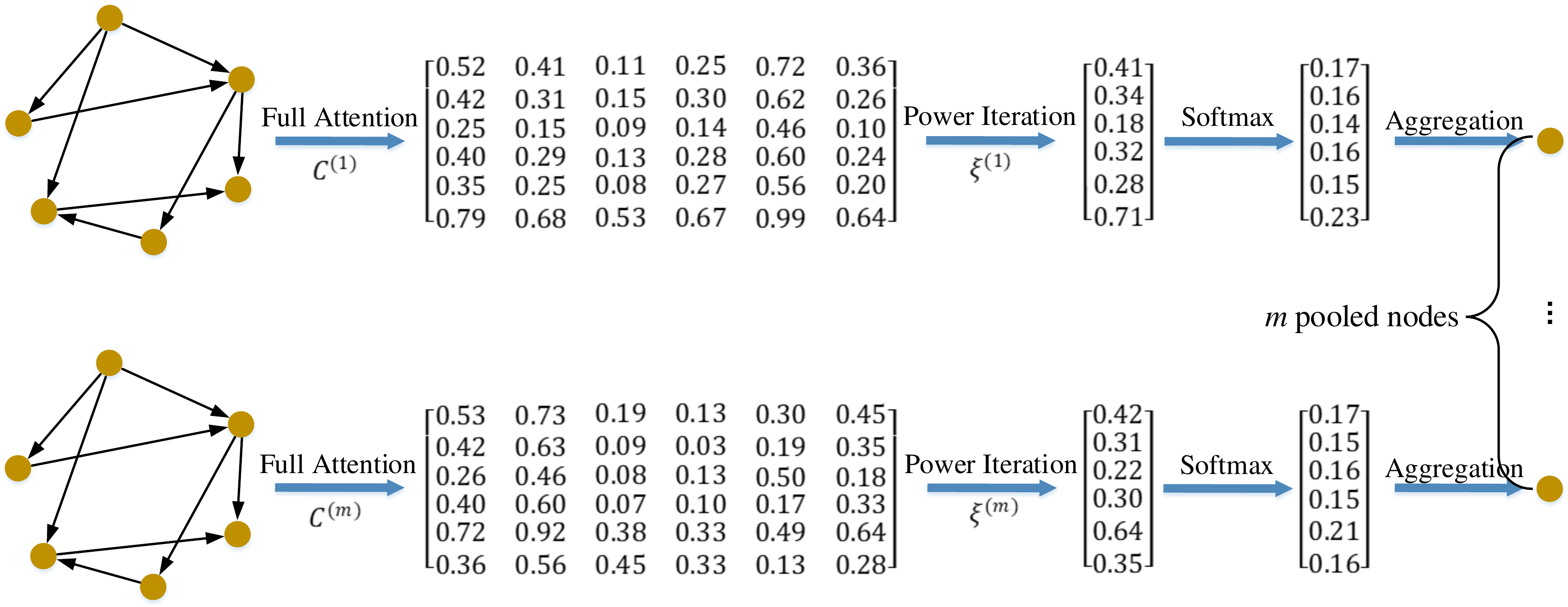}
	\caption{The pipeline of MHFAPool, which consists of $m$ attention branches. For the $k$-th branch, we first derive a full attention score matrix $C^{(k)} \in R^{n \times n}$, then compute its main eigenvector $\xi^{(k)}$ by the power iteration algorithm, and finally obtain the aggregation weights via Softmax operation.
		The purpose of MHFAPool is that the main eigenvector contains the most important information of a matrix.}
	\label{fig_MHFA}
\end{figure}

\textbf{SAGPool-based Pooling.}
SAGPool defines the node scores $Z \in R^{n\times 1}$ with the self-attention mechanism, which are learned by a separate GNN branch:
\begin{equation}
Z = {\rm softmax} \{GNN_{score} (A, H^{(GC)}) \},
\label{eq_pooling2}
\end{equation}
where the output dimension of $GNN_{score}$ is equal to $1$.
In the pooling layer, a separate GNN branch, i.e., $GNN_{score} (\cdot)$ is constructed to learn the node scores via Eq.~(\ref{eq_pooling2}).
In the training stage, $GNN_{score} (\cdot)$ learns the node scores via self-attention mechanism.
Then SAGPool ranks the node with respect to their scores.
To downsample the input graph, SAGPool retains the nodes with top-$m$ scores and discards the remaining nodes.
Thereby, the number of nodes decreases from $n$ to $m$ after the pooling layer.

\textbf{Multi-Head Full Attention Pooling.}
Essentially, DiffPool and SAGPool perform the local node clustering and node selection on graph data, respectively.
While our MHFAPool achieves the multi-branch global node aggregation via the multi-head attention mechanism as shown in Fig.~\ref{fig_MHFA}.
Specifically, for the $k$-th branch, we first compute a full attention coefficient matrix $C^{(k)}=\{ c_{ij}^{(k)} | i,j=1,2,\cdots,n \}$ as follows:
\begin{equation}
c_{ij}^{(k)} = {\rm ReLU} \{[h^{(GC)}_i || h^{(GC)}_j] w^{(k)} \},
\label{eq_att_score}
\end{equation}
where $h^{(GC)}_i$ denotes the node feature of $v_i$ after the graph convolutional layers, and $\cdot || \cdot$ is the concatenation operation.

The coefficient matrix $C^{(k)} \in R^{n \times n}$ is a square matrix that indicates the node cross-correlations.
We aim to obtain a $n$-dimensional aggregation vector for the global node clustering.
Considering that the main eigenvector of a matrix contains the richest information, we employ the power iteration, a fast and differential algorithm to calculate the main eigenvector for matrices, to perform the dimensionality reduction on $C^{(k)}$.
Specifically, the power iteration first randomly initializes a column vector $\xi^{(k)} \in R^{n \times 1}$, then repeats the following steps until convergence:
\begin{equation}
\xi^{(k)} = C^{(k)} \xi^{(k)} ,
\end{equation}
\begin{equation}
\xi^{(k)} = \frac{1}{\Vert \xi^{(k)} \Vert_2} \xi_k .
\end{equation}
After no more than 5 iterations, $\xi^{(k)}$ converges to the main eigenvector of $C^{(k)}$.
Then the feature of the $k$-th pooled node $h^{(P)}_k$ can be derived as follows:
\begin{equation}
h^{(P)}_k = {\rm softmax} \{{\xi^{(k)}}^T \} H^{(GC)}.
\label{eq_mhfa_att}
\end{equation}
If we implement the multi-head structure, for instance, $m$ heads, we could obtain $m$ pooled nodes denoted as $H^{(P)}$.
The detailed steps of MHFAPool are summarized in Algorithm~\ref{algo_MHFA}.

\begin{algorithm}[t]
	\caption{MHFAPool}
	\label{algo_MHFA}
	\LinesNumbered
	\KwIn{Head number $m$, node features $H^{(GC)}$.}
	\KwOut{Pooled node features $H^{(P)}$}
	\For{$k=1:m$}
	{
		$c_{ij}^{(k)}$ $\leftarrow$ ${\rm ReLU} \{[h^{(GC)}_i || h^{(GC)}_j] w^{(k)} \}$ ; \\
		Initialize $\xi^{(k)}$ randomly; \\
		\Repeat{convergence}{
			$\xi^{(k)}$ $\leftarrow$ $C^{(k)}$ $\xi^{(k)}$; \\
			$\xi^{(k)}$ $\leftarrow$ $\frac{1}{\Vert \xi^{(k)} \Vert_2}$ $\xi_k$; \\}
		$h^{(P)}_k$ $\leftarrow$ ${\rm softmax} \{{\xi^{(k)}}^T \} H^{(GC)}$; \\
	}
	$H^{(P)}$ $\leftarrow$ [$h^{(P)}_1$, $h^{(P)}_2$,$\cdots$,$h^{(P)}_m$].
\end{algorithm}

\subsubsection{Readout Layer}
With the pooling layer, the number of nodes in graph is reduced from $n$ to $m$.
To obtain the graph representation $f_{\mathcal{G}}$ with a fixed length, we implement a readout layer as SAGPool~\cite{SAGPool}:
\begin{equation}
f_{\mathcal{G}} = FC (\frac{1}{m} \sum_{i=1}^{m} h^{(P)}_i || \max \limits_{i} \{h^{(P)}_i\} ),
\label{eq_readout}
\end{equation}
where FC is the fully-connected layer, $h^{(P)}_i$ denotes the node feature in $H^{(P)}$.
Given the $p$-th order granularity features, we thus can learn a representation vector $f_{\mathcal{G}}^{(p)}$ with the above-mentioned layers with fully exploiting the temporal and spatial clues.

\subsection{Model Learning}
\label{loss}
As shown in Fig.~\ref{fig_model}, our GPNet contains multiple branches, where each branch can learn a graph representation vector from the input features.
Thus, we obtain the graph representations $f_{\mathcal{G}}^{(p)}$ for $p$ in $\{1,2,4,8\}$.
We then concatenate $f_{\mathcal{G}}^{(p)}$ to construct the final representation vector $f_i$ for the input video sequence $V_i$.

In this paper, we combine the triplet loss and the identity loss to train our GPNet, where the hard triplet mining strategy is employed in the triplet loss:
\begin{equation}
L_{triplet} = \sum_{i=1}^{N} \max(0, \delta + \max \limits_{\substack{y_i=y_j \\   j=1,2,\cdots,N}} \{ \| f_i - f_j \|_2 \} -\min \limits_{\substack{y_i\neq y_k \\   k=1,2,\cdots,N}} \{ \| f_i - f_k \|_2 \}),
\label{eq_triplet_loss}
\end{equation}
\begin{equation}
L_{identity} = -\frac{1}{N} \sum_{i=1}^{N} \log(\frac{\exp(z_i)}{\sum_{i=1}^{N} \exp(z_i)}) .
\label{eq_identity_loss}
\end{equation}
In Eq.~(\ref{eq_triplet_loss}), $\delta$ denotes the pre-defined margin and $N$ indicates the training batch size.
In Eq.~(\ref{eq_identity_loss}), $z_i=FC(f_i)$ denotes the classification logits.
In the inference stage, the representation vector $f_i$ is utilized for the pedestrian retrieval.

\section{Experiments}
\label{experiments}
In this section, we first introduce the experimental setup, including the datasets, evaluation metrics and the experimental implementations, then analyze the validity of each module in our GPNet, last compare our GPNet with existing state-of-the-art methods in Section~\ref{Comparison}.

\subsection{Experimental Setup}
\label{ES}
\textbf{Datasets.}
In this paper, we test our GPNet on four video datasets: MARS~\cite{Mars}, DukeMTMC-VideoReID~\cite{duke1,duke2}, iLIDS-VID~\cite{ilids} and PRID-2011~\cite{prid}. 
MARS~\cite{Mars} is currently the largest video-based person ReID dataset, which contains $1,261$ identities ($625$ identities for training and $636$ identities for testing), where each identity has an average of $13.9$ sequences.
DukeMTMC-VideoReID~\cite{duke1,duke2} is another large video-based person ReID dataset with $1,812$ identities that are divided into $408$, $702$ and $702$ identities for distraction, training, and testing, respectively.
iLIDS-VID~\cite{ilids} contains $600$ video sequences from $300$ identities, where the average length of the sequences is $45$.
PRID-2011~\cite{prid} contains $749$ pedestrians captured in the uncrowded background, and the average length of sequences are about $100$.

\textbf{Evaluation metrics.}
With respect to MARS and DukeMTMC-VideoReID, we report both the Cumulative Match Characteristic (CMC) and mean Average Precision (mAP).
Following the experimental setting in ~\cite{STGCN,MGH}, we only present the CMC values on iLIDS-VID for comparison.

\textbf{Implementations.}
In this paper, the ResNet50~\cite{ResNet} is taken as the backbone, and the non-local blocks~\cite{nonlocal} are employed in the network as MGH~\cite{MGH}.
We resize the pedestrian images in the video sequences into $256\times 128$.
In the training stage, each batch is composed of $32$ video sequences from $8$ identities, and the length of each video sequence $T$ is set as $8$.
The length of $p$-th order granularity features is equal to $2048$ for $p=1$, and equal to $1024$ for $p=\{2,4,8\}$, so that the representation vector of the input video sequence is $5120$-dimensional.
In Eq.~(\ref{eq_edge}), we set $\delta_T=1$ and the number of neighbors $K=2$.
We choose to retain one quarter nodes after the pooling layer, in other words, the number of pooled nodes after the pooling layer is $2,4,8,16$ for $p=\{1,2,4,8\}$, respectively.
We train our GPNet $800$ epochs with a warm-up strategy; we utilize the Adam~\cite{Adam} optimizer with $0.0005$ weight decay to train our network; the initial learning rate is set to $0.0003$ and it decays by 0.1 every $100$ epochs.

\begin{table*}[ht] 
	\footnotesize
	\centering
	\caption{Performance of our GPNet with different convolution and pooling methods on MARS~\cite{Mars} and DukeMTMC-VideoReID~\cite{duke1,duke2}, where \textit{Spatial Conv} and \textit{Spectral Conv} denote the spatial-based convolution and the spectral-based convolution, respectively. The best performance is marked in \textbf{bold} and \textcolor{red}{red}.}
	\begin{tabular}{c|c|cccc|cccc}
		\hline
		\multirow{2}{*}{GC layer}                  & \multirow{2}{*}{Pooling layer} & \multicolumn{4}{c|}{MARS}     & \multicolumn{4}{c}{DukeMTMC-VideoReID}      \\ \cline{3-10} 
		&                                & mAP  & Rank1 & Rank5 & Rank20 & mAP  & Rank1 & Rank5 & Rank20 \\ \hline
		\multirow{5}{*}{Spatial Conv} & Mean pooling                   & 83.5 & 87.3  & 96.3  & 98.2   & 95.3 & 95.4  & 99.2  & 99.6   \\
		& Max pooling                   & 82.6 & 87.1  & 95.9  & 97.6   & 95.2 & 95.1  & 99.0  & 99.6   \\
		& DiffPool~\cite{DiffPool}                   & 84.7 & 89.6  & \textbf{\textcolor{red}{96.8}}  & 98.7  & 96.0 & 96.3  & 99.3  & 99.8  \\
		& SAGPool~\cite{SAGPool}                & 84.1 & 88.2  & 96.6  & 98.7  & 96.1 & 96.7  & \textbf{\textcolor{red}{99.6}}  & \textbf{\textcolor{red}{99.9}}    \\ 
		& MHFAPool                & \textbf{\textcolor{red}{84.9}} & \textbf{\textcolor{red}{89.9}}  & \textbf{\textcolor{red}{96.8}}  & \textbf{\textcolor{red}{98.8}}  & 96.1 & 96.9  & \textbf{\textcolor{red}{99.6}}  & \textbf{\textcolor{red}{99.9}}    \\ \hline
		\multirow{5}{*}{Spectral Conv} & Mean pooling                   & 83.2 & 87.1  & 96.1  & 98.2   & 95.5 & 96.2  & 99.0  & 99.4   \\
		& Max pooling                   & 82.6 & 87.0  & 95.9  & 98.0   & 95.8 & 96.3  & 99.1  & 99.3   \\
		& DiffPool~\cite{DiffPool}                   & 84.5 & 88.8  & 96.6  & 98.5  & 96.2 & 97.2  & 99.5  & 99.8   \\
		& SAGPool~\cite{SAGPool}                 & 83.9 & 88.0  & 96.7  & 98.4   & \textbf{\textcolor{red}{96.3}} & 97.2  & \textbf{\textcolor{red}{99.6}}  & \textbf{\textcolor{red}{99.8}}  \\
		& MHFAPool      & 84.7 & 89.2  & 96.6  & 98.6   & \textbf{\textcolor{red}{96.3}} & \textbf{\textcolor{red}{97.3}}  & \textbf{\textcolor{red}{99.6}}  & \textbf{\textcolor{red}{99.8}}    \\ \hline
	\end{tabular}
	\label{table_components}
\end{table*}

\begin{table}[t]
	\footnotesize
	\centering
	\caption{Performance of our GPNet with different feature granularities on MARS~\cite{Mars}, where p=1 means that we only adopt the global features. The best performance is marked in \textbf{bold} and \textcolor{red}{red}.}
	\begin{tabular}{c|c|cccc}
		\hline
		\multirow{2}{*}{GC layer}           & \multirow{2}{*}{$p$} & \multicolumn{4}{c}{Mars}      \\ \cline{3-6} 
		&                              & mAP  & Rank1 & Rank5 & Rank20 \\ \hline
		\multirow{4}{*}{Spatial Conv} & $\{1\}$                      & 83.5 & 87.9  & 96.6  & 98.6   \\
		& $\{1,2\}$                    & 84.0 & 88.1  & 96.5  & 98.4   \\
		& $\{1,2,4\}$                  & 84.3 & 88.5  & 96.6  & 98.6   \\
		& $\{1,2,4,8\}$                & \textbf{\textcolor{red}{84.9}} & \textbf{\textcolor{red}{89.9}}  & \textbf{\textcolor{red}{96.8}}  & \textbf{\textcolor{red}{98.8}}   \\ \hline
		\multirow{4}{*}{Spectral Conv} & $\{1\}$                      & 83.5 & 87.3  & 96.5  & 98.3   \\
		& $\{1,2\}$                    & 84.3 & 88.6  & 96.6  & 98.3   \\
		& $\{1,2,4\}$                  & 84.4 & 88.6  & 96.7  & 98.3   \\
		& $\{1,2,4,8\}$                & 84.7 & 89.2  & 96.6  & 98.6   \\ \hline
	\end{tabular}
	\label{table_granularity}
\end{table}

	\subsection{Ablation Studies}
\label{Ablation_Studies}

\subsubsection{Analysis on GPNet Components}
In this section, we explore the performance of our GPNet with different types of graph convolutions and pooling methods on MARS~\cite{Mars} and DukeMTMC-VideoReID~\cite{duke1,duke2}.
Specifically, we attempt the spatial-based graph convolution and spectral-based convolution for the GC layers, meanwhile we implement DiffPool~\cite{DiffPool}, SAGPool~\cite{SAGPool} and MHFAPool for the pooling layer in GPNet.
Otherwise, we compare the mean and max pooling with the above-mentioned pooling methods, where the mean/max pooling takes the average/max value of node features as the graph representation.

We present the experimental results in Table~\ref{table_components}, from which we can draw the following findings:
the spatial-based convolution performs better than the spectral-based convolution on MARS, while the spectral-based convolution achieves better performance on DukeMTMC-VideoReID;
DiffPool and SAGPool obtain similar results on DukeMTMC-VideoReID, while DiffPool is superior than SAGPool on Mars;
DiffPool and SAGPool outperform the mean and max pooling on both MARS and DukeMTMC-VideoReID, which verifies the effectiveness of the graph pooling layer;
moreover, MHFAPool outperforms DiffPool and SAGPool, which demonstrates the validity of our multi-head attention and power iteration.

\begin{table}[t]
	\footnotesize
	\centering
	\caption{Performance of our GPNet with different numbers of retained nodes after pooling layer on MARS~\cite{Mars}, where ratio is equal to the number of pooled nodes divided by that of total nodes. The best performance is marked in \textbf{bold} and \textcolor{red}{red}.}
	\begin{tabular}{c|c|cccc}
		\hline
		\multirow{2}{*}{GC layer}           & \multirow{2}{*}{Ratio} & \multicolumn{4}{c}{Mars}      \\ \cline{3-6} 
		&                              & mAP  & Rank1 & Rank5 & Rank20 \\ \hline
		\multirow{3}{*}{Spatial Conv} 
		& 1/4                    & \textbf{\textcolor{red}{84.9}} & \textbf{\textcolor{red}{89.9}}  & \textbf{\textcolor{red}{96.8}}  & \textbf{\textcolor{red}{98.8}}   \\
		& 1/3                  & 84.7 & 89.6  & 96.6  & 98.6   \\
		& 1/2                & 84.8 & 89.8  & \textbf{\textcolor{red}{96.8}}  & 98.7   \\ \hline
		\multirow{3}{*}{Spectral Conv} 
		& 1/4                    & 84.7 & 89.2  & 96.6  & 98.6   \\
		& 1/3                  & 84.6 & 88.9  & 96.7  & 98.3   \\
		& 1/2                & 84.6 & 89.0  & 96.6  & 98.4   \\ \hline
	\end{tabular}
	\label{table_npooled}
\end{table}

\subsubsection{Impact of the Granularity}
In this section, we explore the effectiveness of the multi-granularity features on Mars~\cite{Mars}.
Specifically, we conduct our experiments with the features of different granularities: $p=\{1\}$, $p=\{1,2\}$, $p=\{1,2,4\}$, $p=\{1,2,4,8\}$, where $p=\{1\}$ means that we only employ the global features to construct the global-branch graph.
We test both spatial-based and spectral-based convolution for the GC layer, and we implement the pooling layer with MHFAPool.
We report the results in Table~\ref{table_granularity}.
As can be seen, the case of $p=\{1\}$ presents the worst performance, which results from the neglect of the spatial clues.
Otherwise, we conclude that the fine-gained features can effectively improve the performance under different graph convolutions: the GPNet with $p=\{1,2,4,8\}$ achieves the best results under both spatial-based convolution and spectral-based convolution.

\subsubsection{Analysis on the Number of Retained Nodes}
In this section, we explore the impact of $n_{pooled}$, i.e., the number of retained nodes.
We test both spatial-based and spectral-based convolution for the GC layer, and we implement the pooling layer with MHFAPool.
We report the results in Table~\ref{table_npooled}, where the numbers denote the keep ratio, i.e., the number of pooled nodes divided by that of total nodes.
As can be seen, the performance is slightly affected by this parameter.
Conclusion could be drawn that most nodes are not important for retrieval accuracy, while retaining fewer nodes can reduce storage and speed up inference.

\begin{table}[t]
	\footnotesize 
	\centering
	\caption{Comparison of different adjacency methods on MARS~\cite{Mars}. SA denotes the self-attention adjacency. The best performance is marked in \textbf{bold} and \textcolor{red}{red}.}
	\begin{tabular}{c|c|cccc}
		\hline
		\multirow{2}{*}{GC layer}           & \multirow{2}{*}{Adjacency} & \multicolumn{4}{c}{Mars}      \\ \cline{3-6} 
		&                               & mAP  & Rank1 & Rank5 & Rank20 \\ \hline
		\multirow{4}{*}{Spatial Conv} & SA                     & 82.5 & 87.0  & 96.1  & 98.3   \\
		& TNA                           & 84.3 & 88.9  & 96.4  & 98.5   \\
		& ENA                           & 84.7 & 88.4  & 96.7  & 98.5   \\
		& DNA                          & \textbf{\textcolor{red}{84.9}} & \textbf{\textcolor{red}{89.9}}  & \textbf{\textcolor{red}{96.8}}  & \textbf{\textcolor{red}{98.8}}   \\ \hline
		\multirow{4}{*}{Spectral Conv} & SA                     & 82.6 & 86.8  & 95.9  & 98.4   \\
		& TNA                           & 84.4 & 88.0  & 96.6  & 98.4   \\
		& ENA                           & 84.4 & 87.8  & 96.5  & 98.7   \\
		& DNA                           & 84.7 & 89.2  & 96.6  & 98.5   \\ \hline
	\end{tabular}
	\label{table_adjacency}
\end{table}

\subsubsection{Comparison with Other Adjacency Methods}
In Section~\ref{GPNet}, we propose the dual-neighborhood aggregation (DNA) mechanism to construct the connections between the nodes.
In this section, we compare the DNA mechanism with the temporal-neighborhood aggregation (TNA) and the Euclidean-neighborhood aggregation (ENA) mechanism, where TNA mechanism constructs the edges between the temporal-neighborhood nodes, and ENA mechanism constructs the edges between the Euclidean-neighborhood nodes.
Otherwise, we explore the performance of the self-attention adjacency:
\begin{equation}
A_{ij} = {\rm softmax}_j \{ {\rm LeakyReLU} ({\vec{a}}^T [f_i || f_j]) \},
\label{eq_self_attention}
\end{equation}
where $\vec{a}$ is a learnable vector.
Eq.~(\ref{eq_self_attention}) defines the fully connected graph with the self-attention mechanism. 
We test both spatial-based and spectral-based convolution for the GC layer, and we implement the pooling layer with MHFAPool.
We present the experimental results in Table~\ref{table_adjacency}.
As can be seen, the self-attention adjacency reports the worst performance, and our DNA mechanism outperforms both ENA and TNA mechanism.

\begin{figure}[h]
	\centering
	\subfigure[Spatial-based convolution]{\includegraphics[width=0.3\textwidth]{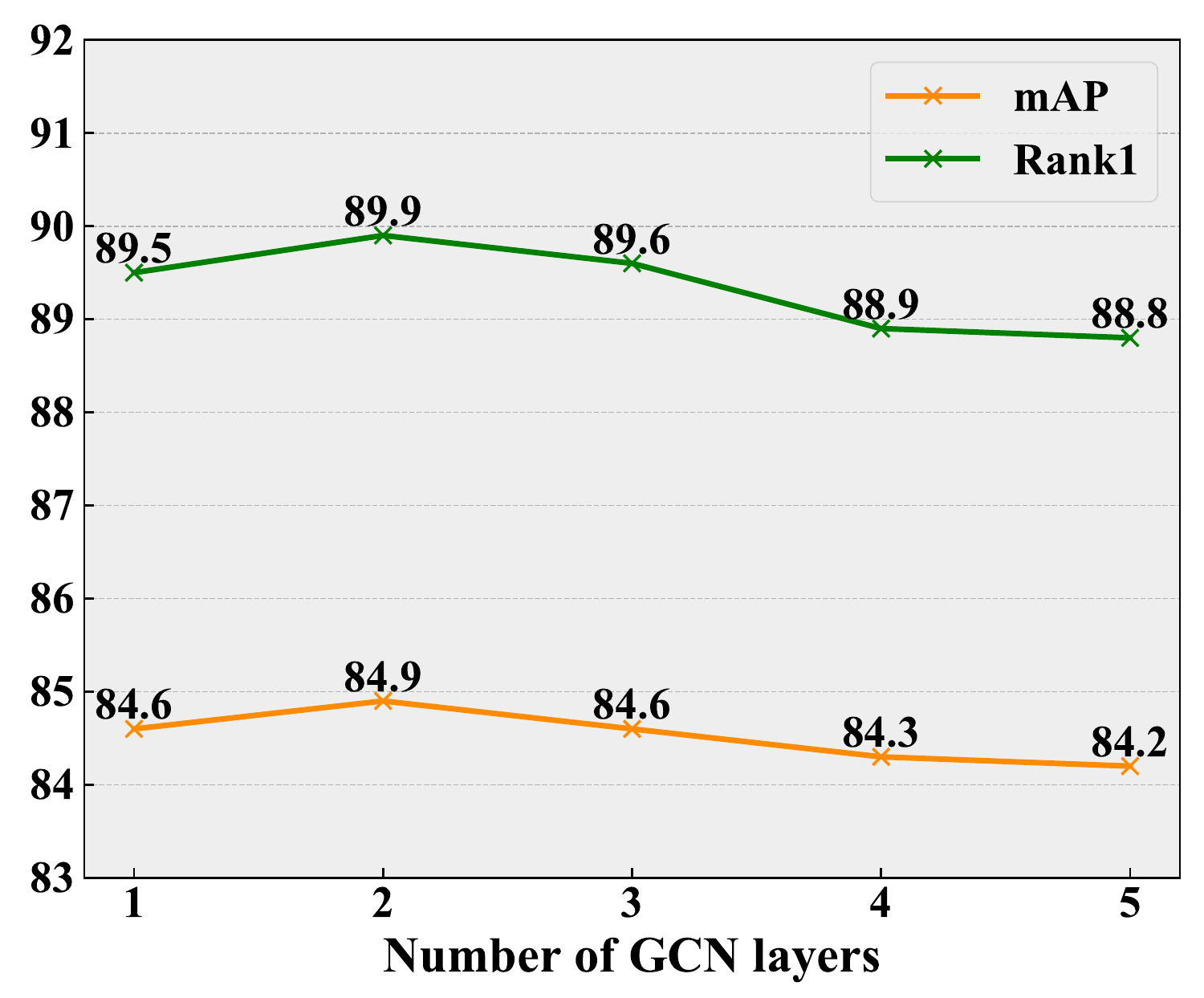}}
	\subfigure[Spectral-based convolution]{\includegraphics[width=0.3\textwidth]{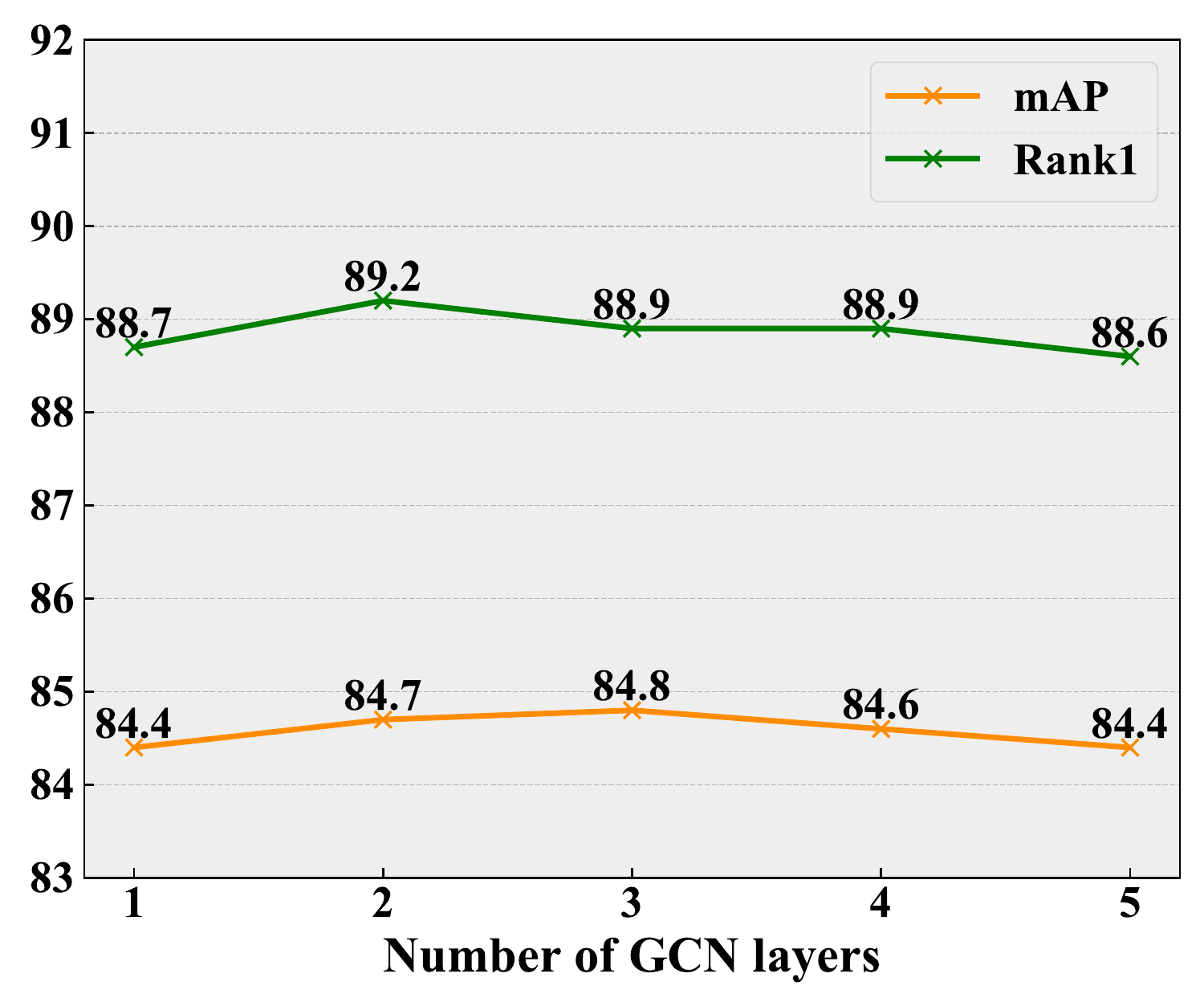}}
	\caption{Analysis on the number of GC layers.
		In (a), we found that the performance declines when the layer number exceeds 3.
		In (b), the retrieval accuracy remains stable as the number of GC layers increase, which benefits from the shortcut structure.}
	\label{fig_gcn_layer}
\end{figure}

\begin{figure}[h]
	\centering
	\subfigure[Spatial-based convolution]{\includegraphics[width=0.3\textwidth]{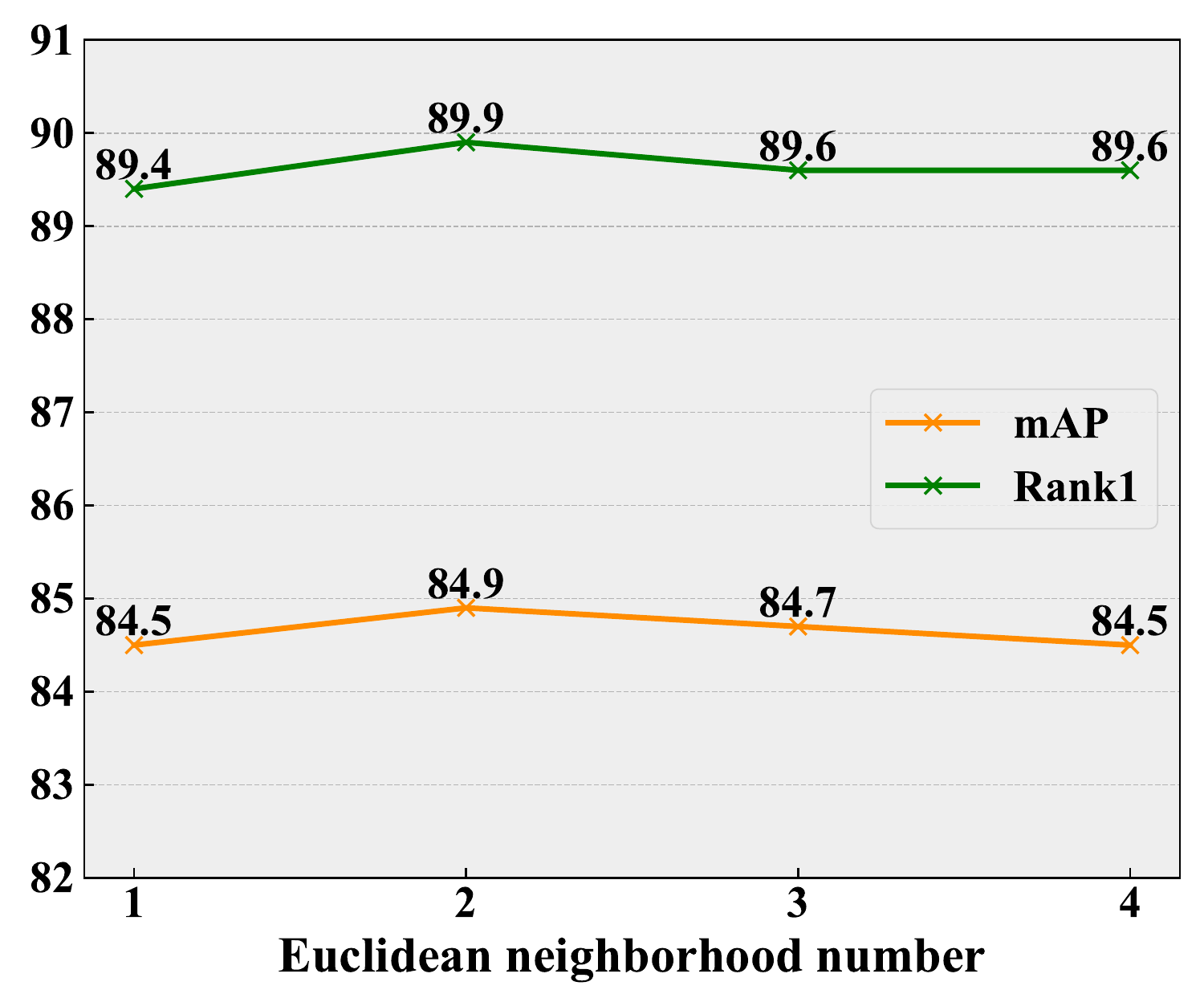}}
	\subfigure[Spectral-based convolution]{\includegraphics[width=0.3\textwidth]{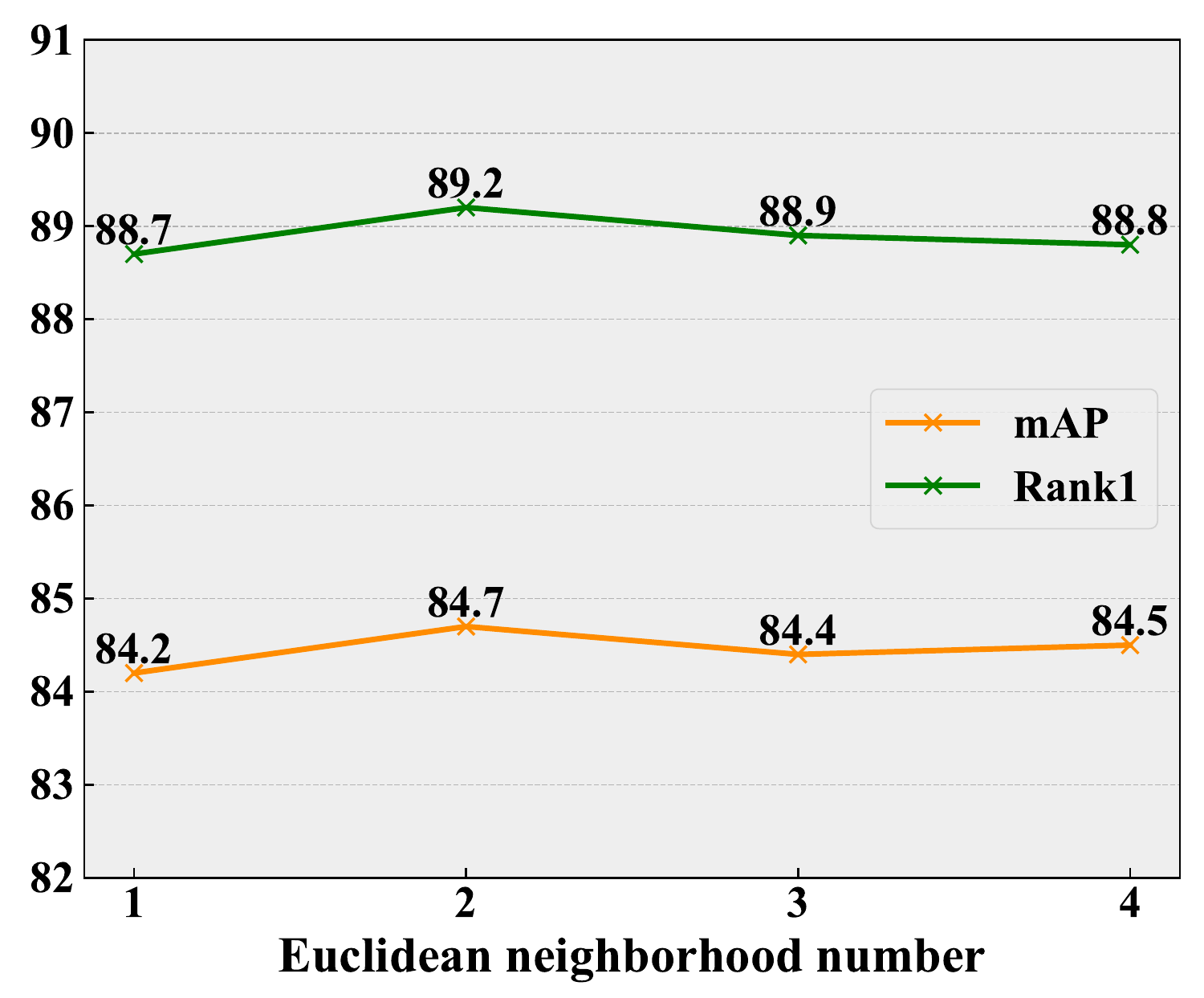}}
	\caption{Analysis on the Euclidean neighborhood number k.
		For both spatial-based and spectral-based convolution, the GPNet achieves the best performance when k=2.
	}
	\label{fig_gcn_k}
\end{figure}

\begin{table}[t] 
	\footnotesize
	\centering
	\caption{Performance of our GPNet under different sequence lengths T on MARS~\cite{Mars}. The best performance is marked in \textbf{bold} and \textcolor{red}{red}.}
	\begin{tabular}{c|c|cccc}
		\hline
		\multirow{2}{*}{GC layer}      & \multirow{2}{*}{Length} & \multicolumn{4}{c}{Mars}      \\ \cline{3-6} 
		&                         & mAP  & Rank1 & Rank5 & Rank20 \\ \hline
		\multirow{4}{*}{Spatial Conv}  & T=4                     & 84.3 & 89.8  & 96.6  & 98.5   \\
		& T=6                     & 84.4 & 89.5  & \textbf{\textcolor{red}{96.8}}  & 98.4   \\
		& T=8                     & 84.9 & 89.9  & \textbf{\textcolor{red}{96.8}}  & \textbf{\textcolor{red}{98.8}}   \\
		& T=10                     & \textbf{\textcolor{red}{85.1}} & \textbf{\textcolor{red}{90.2}}  & 96.7  & \textbf{\textcolor{red}{98.8}}   \\ \hline
		\multirow{4}{*}{Spectral Conv} & T=4      & 84.1 & 88.0 & 96.6  & 98.4   \\
		& T=6                     & 84.3 & 88.7  & 96.7  & 98.4   \\
		& T=8                     & 84.7 & 89.2  & 96.6  & 98.5   \\ 
		& T=10                     & 84.8 & 89.3  & 96.6  & 98.6   \\ \hline
	\end{tabular}
	\label{table_length}
\end{table}

\subsubsection{Analysis on the Number of GC Layers}
The number of graph convolutional layers is a trade-off.
With more GC layers, each node could aggregate more neighborhood information.
However, the multi-layer GNNs would also result in the over-smoothing~\cite{smoothing}.
In this section, we verify the performance of the GPNet with different numbers of the graph convolutional layers on MARS~\cite{Mars}.
We test both spatial-based and spectral-based convolution for the GC layer, and we implement the pooling layer with MHFAPool.

As shown in in Fig.~\ref{fig_gcn_layer}.
For the spatial-based convolution, the the two-layer architecture achieves the best performance, and when the number of layers exceeds 2, the performance will continue to decrease.
For the spectral-based convolution, thanks to the shortcut connection of Eq.~(\ref{eq_pooling2}), the retrieval accuracy remains stable as the number of GC layers increases.

\subsubsection{Analysis on the Model Parameters}
In this section, we test the performance of the GPNet under different sequence lengths $T$ and Euclidean neighborhood numbers $k$.
Specifically, we conduct the experiments under the setting $T=\{4,6,8\}$ and $k=\{1,2,3,4\}$.
We test both spatial-based and spectral-based convolution for the GC layer, and we implement the pooling layer with MHFAPool.
Their experimental results on MARS~\cite{Mars} are presented in Table~\ref{table_length} and Fig.~\ref{fig_gcn_k}, respectively.
From Table~\ref{table_length} we conclude that enlarging $T$ could improve the performance for both spatial-based and spectral-based convolution;
Fig.~\ref{fig_gcn_k} indicates that $k$ slightly affects the experimental accuracy, and GPNet achieves the best performance when $k=2$.

\begin{table}[h]
	\footnotesize
	\centering
	\caption{Comparison with the state-of-the-arts on MARS~\cite{Mars}. The three best scores are indicated in \textcolor{red}{red}, \textcolor{blue}{blue}, \textcolor{green}{green}, respectively.}
	\begin{tabular}{c|c|cccc}
		\hline
		Method & Venue & mAP  & Rank1 & Rank5 & Rank20 \\ \hline
		ADFD~\cite{ADFD}     & CVPR 2019  & 78.2   & 87.0    & 95.4    & \textcolor{blue}{98.7}       \\
		VRSTC~\cite{VRSTC}    & CVPR 2019  & 82.3   & 88.5    & 96.5    & 97.4       \\
		GLTR~\cite{GLTR}     & ICCV 2019  & 78.5   & 87.0    & 95.8    & 98.2       \\
		TCLNet~\cite{TCLNet}   & ECCV 2020  & 83.0   & 88.8    & -       & -          \\
		AFA~\cite{AFA}	     & ECCV 2020  & 82.9   & \textcolor{blue}{90.2}    & \textcolor{green}{96.6}    & -          \\
		STGCN~\cite{STGCN}    & CVPR 2020  & 83.7   & 89.9    & 96.4    & 98.2       \\
		MGH~\cite{MGH}	     & CVPR 2020  & \textcolor{blue}{85.8}   & 90.0    & \textcolor{blue}{96.7}    & \textcolor{green}{98.5}       \\ 
		SSN3D~\cite{SSN3D}   & AAAI 2021  & \textcolor{red}{86.2}   & \textcolor{green}{90.1}    & \textcolor{green}{96.6}       & 98.0          \\ 
		GRL~\cite{GRL}   & CVPR 2021  & 84.8   & \textcolor{red}{91.0}    & \textcolor{blue}{96.7}       & 98.4          \\
		BiCnet~\cite{BiCnet}   & CVPR 2021  & \textcolor{green}{86.0}   & \textcolor{blue}{90.2}    & -       & -          \\  \hline
		GPNet    & -     & 85.1 & \textcolor{blue}{90.2}  & \textcolor{red}{96.8}  & \textcolor{red}{98.8}   \\ \hline
	\end{tabular}
	\label{table_mars}
\end{table}

\begin{table}[h]
	\footnotesize
	\centering
	\caption{Comparison with the state-of-the-arts on DukeMTMC-VideoReID~\cite{duke1,duke2}. The three best scores are indicated in \textcolor{red}{red}, \textcolor{blue}{blue}, \textcolor{green}{green}, respectively.}
	\begin{tabular}{c|c|cccc}
		\hline
		Method & Venue & mAP  & Rank1 & Rank5 & Rank20 \\ \hline
		VRSTC~\cite{VRSTC}    & CVPR 2019  & 93.5   & 95.0    & 99.1    & -       \\
		COSAM~\cite{COSAM}    & ICCV 2019  & 94.1   & 95.4    & \textcolor{green}{99.3}    & \textcolor{blue}{99.8}       \\
		GLTR~\cite{GLTR}     & ICCV 2019  & 93.7   & 96.3    & \textcolor{green}{99.3}    & \textcolor{green}{99.7}       \\
		TCLNet~\cite{TCLNet}   & ECCV 2020  & \textcolor{blue}{96.2}   & \textcolor{green}{96.9}    & -       & -          \\
		AFA~\cite{AFA}	     & ECCV 2020  & 95.4   & \textcolor{blue}{97.2}    & \textcolor{blue}{99.4}    & \textcolor{red}{99.9}          \\
		AP3D~\cite{3DCNN1}     & ECCV 2020  &  \textcolor{green}{96.1}   & \textcolor{blue}{97.2}    & -       & -          \\
		STGCN~\cite{STGCN}    & CVPR 2020  & 95.7   & \textcolor{red}{97.3}    & \textcolor{green}{99.3}    & \textcolor{blue}{99.8}       \\
		SSN3D~\cite{SSN3D}    & AAAI 2021  & \textcolor{red}{96.3}   & 96.8    & -    & -       \\ 
		BiCnet~\cite{BiCnet}   & CVPR 2021  & \textcolor{green}{86.0}   & \textcolor{blue}{90.2}    & -       & -          \\  \hline
		GPNet    & -     & \textcolor{green}{96.1} & 96.3  & \textcolor{red}{99.6}  & \textcolor{blue}{99.8}   \\ \hline
	\end{tabular}
	\label{table_duke}
\end{table}

\begin{table*}[h]
	\footnotesize
	\centering
	\caption{Comparison with the state-of-the-arts on iLIDS-VID~\cite{ilids} and PRID-2011~\cite{prid}. The three best scores are indicated in \textcolor{red}{red}, \textcolor{blue}{blue}, \textcolor{green}{green}, respectively.}
	\begin{tabular}{c|c|cccccc}
		\hline
		Method  & \multirow{2}{*}{Venue} & \multicolumn{3}{c|}{iLIDS-VID}              & \multicolumn{3}{c}{PRID-2011} \\ \cline{3-8} 
		&                        & Rank1 & Rank5 & \multicolumn{1}{c|}{Rank20} & Rank1    & Rank5   & Rank20   \\ \hline
		ADFD~\cite{ADFD}    & CVPR 2019              & 86.3  & 97.4  & \multicolumn{1}{c|}{\textcolor{green}{99.7}}   & 93.9         & 99.5        & \textcolor{red}{100}         \\
		VRSTC~\cite{VRSTC}   & CVPR 2019              & 83.4  & 95.5  & \multicolumn{1}{c|}{99.5}   & -         & -        & -         \\
		GLTR~\cite{GLTR}    & ICCV 2019              & 86.0  & \textcolor{green}{98.0}  & \multicolumn{1}{c|}{-}      & \textcolor{green}{95.5}         & \textcolor{red}{100}        & -         \\
		TCLNet~\cite{TCLNet}  & ECCV 2020              & 86.6  & -     & \multicolumn{1}{c|}{-}      & -         & -        & -         \\
		AFA~\cite{AFA}     & ECCV 2020              & 88.5  & 96.8  & \multicolumn{1}{c|}{\textcolor{green}{99.7}}   & -         & -        & -         \\
		MGH~\cite{MGH}     & CVPR 2020              & 85.6  & 97.1  & \multicolumn{1}{c|}{99.5}   & 94.8         & 99.3        & \textcolor{red}{100}         \\
		SSN3D~\cite{SSN3D}  & AAAI 2021              & \textcolor{blue}{88.9}  & 97.3     &  \multicolumn{1}{c|}{98.8}      & -         & -        & -         \\
		GRL~\cite{GRL}  & CVPR 2021              & \textcolor{red}{90.4}  & \textcolor{blue}{98.3}     &  \multicolumn{1}{c|}{\textcolor{blue}{99.8}}      & \textcolor{red}{96.2}         & \textcolor{green}{99.7}        & \textcolor{red}{100}         \\
		\hline
		GPNet   & -                      & \textcolor{green}{88.8}  & \textcolor{red}{98.5}  & \multicolumn{1}{c|}{\textcolor{red}{100}}  & \textcolor{blue}{96.1}         & \textcolor{blue}{99.8}        & \textcolor{red}{100}        \\ \hline
	\end{tabular}
	\label{table_ildis}
\end{table*}

\subsection{Comparison and Visualization}
\label{Comparison}

\subsubsection{Comparison with state-of-the-arts}
\label{SOTA}
In this section, we compare our GPNet with the current state-of-the-arts on three widely-used datasets, i.e. MARS~\cite{Mars}, DukeMTMC-VideoReID~\cite{duke1,duke2}, iLIDS-VID~\cite{ilids} and PRID-2011~\cite{prid}.
The comparison results are presented in Table~\ref{table_mars}, Table~\ref{table_duke} and Table~\ref{table_ildis}, respectively.
On MARS~\cite{Mars}, we set the sequence length of the input video as $10$; we implement the graph convolutional layer and graph pooling layer with the spatial-based convolution and MHFAPool, respectively.
In Table~\ref{table_mars}, the mAP of our GPNet is larger than that of STGCN~\cite{STGCN} by 1.4\%, while we also outperform the STGCN on the metrics rank5 and rank20.
On DukeMTMC-VideoReID~\cite{duke1,duke2}, we set the sequence length of the input video as $8$; we implement the graph convolutional layer and graph pooling layer with the spectral-based convolution and MHFAPool, respectively.
On PRID-2011~\cite{prid} and iLIDS-VID~\cite{ilids}, we set the sequence length of the input video as $8$, and the other settings are the same as MARS~\cite{Mars}.
By Table~\ref{table_duke} and Table~\ref{table_ildis}, our approach achieves the competitive results on DukeMTMC-VideoReID, iLIDS-VID and PRID-2011~\cite{prid}, for example, our GPNet achieves 96.3\% mAP and 97.3\% rank1 on DukeMTMC-VideoReID, which denotes the best performance.

\subsubsection{Visualization}
\begin{figure*}[t]
	\centering
	\includegraphics[width=0.88\textwidth]{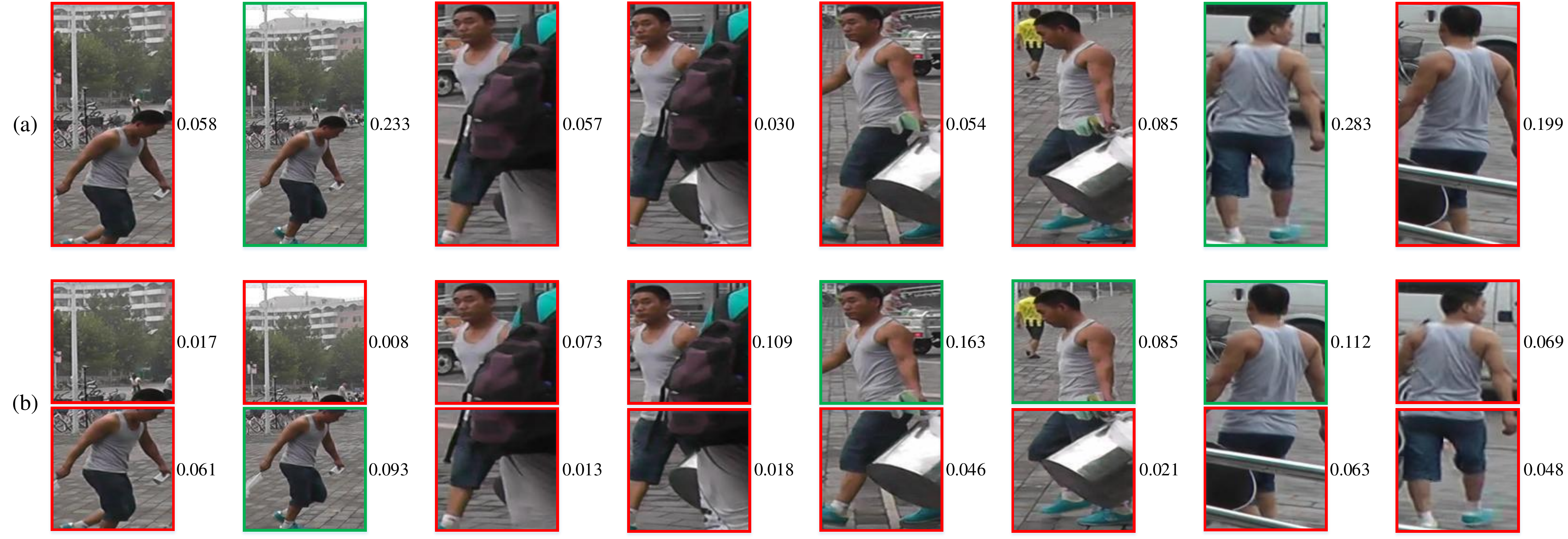}
	\caption{Visualization of the node scores learned by SAGPool~\cite{SAGPool}.
		Images or patches outlined in the green and red boxes indicate the retained nodes and discarded nodes, respectively.}
	\label{fig_vis}
\end{figure*}

In this section, we first visualize the node scores learned by SAGPool~\cite{SAGPool} on a video sequence with 8 frames.
We present the node scores of the global features and 2-nd order granularity features in Fig.~\ref{fig_vis}.
The nodes with top-2 scores and top-4 scores are retained for the global features and 2-nd order granularity features, respectively.
From Fig.~\ref{fig_vis}, we conclude that 1) features containing the occlusion would achieve the low scores; 2) features of the upper body are more important than the lower body.

\section{Conclusions}
\label{conclusion}
To capture the temporal and spatial clues of the video-based ReID, we proposed the GPNet for the multi-granularity feature aggregation.
We first formulated the input graphs with the multi-granularity features, where the edges were constructed for the both temporal and Euclidean neighborhood nodes;
we then implemented the graph convolutional layers for the node neighborhood aggregation.
To obtain the feature representation of the video sequence, we further adopted the graph pooling methods to decrease the number of nodes in graph, which was implemented by a multi-head full attention pooling method with fully considering the global cross-attention.
Finally, we implemented the readout layer to obtain the fix-length representation of the pooled graph.
We concatenated the multi-granularity graph representations as the video representation vector for the pedestrian retrieval.
The experimental results on MARS, DukeMTMC-VideoReID and iLIDS-VID demonstrated the superiority of our GPNet.

\section*{Acknowledgment}
 This research is supported by the National Natural Science
 Foundation of China (Grant No.62172126 and Grant No.62106063), by the Shenzhen Research Council (Grant No. JCYJ20210324120202006), by the Shenzhen College Stability Support Plan (Grant GXWD20201230155427003-20200824113231001).

\bibliography{mybibfile}

\end{document}